\documentclass{article}
\usepackage{graphics} 
\usepackage{graphicx} 
\usepackage{hyperref} 
\usepackage{subcaption}
\usepackage{multirow} 
\usepackage{tabularx} 
\usepackage{color} 
\usepackage{url}
\usepackage{textcomp} 
\usepackage{amsmath} 
\usepackage{amssymb} 
\usepackage{amsfonts} 
\usepackage[margin=1in]{geometry}
\usepackage[numbers]{natbib}
\usepackage[linesnumbered,ruled]{algorithm2e}

\definecolor{color-1}{rgb}{1,1,1}

\title{Automation Strategies for Unconstrained Crossword Puzzle Generation}

\author {Charu Agarwal\footnote{Charu Agarwal is a final year  undergraduate student of Computer Science \& Engineering at IIT Dharwad.}\\ \small{Department of Computer Science \& Engineering,}\\ \small {Indian Institute of Technology Dharwad} \\ \small{Dharwad, India - 580011.}
\and
Rushikesh K. Joshi \footnote{Rushikesh K. Joshi is Professor of Computer Science \& Engineering at IIT Bombay. Email for Correspondence: rkj@iitb.ac.in} \\ \small{Department of Computer Science \& Engineering,} \\ \small{Indian Institute of Technology Bombay}\\ \small{Mumbai, India - 400076.}}

\begin{document}
\date{}
\maketitle
\begin{abstract}

An unconstrained crossword puzzle is a generalization of the constrained crossword problem. In this problem, only the word vocabulary, and optionally the grid dimensions are known. Hence, it not only requires the algorithm to determine the word locations, but it also needs to come up with the grid geometry. This paper discusses algorithmic strategies for automatic crossword puzzle generation in such an unconstrained setting. The strategies proposed cover the tasks of selection of words from a given vocabulary, selection of grid sizes, grid resizing and adjustments, metrics for word fitting, back-tracking techniques, and also clue generation. The strategies have been formulated based on a study of the effect of word sequence permutation order on grid fitting. An end-to-end algorithm that combines these strategies is presented, and its performance is analyzed. The techniques have been found to be successful in quickly producing well-packed puzzles of even large sizes. Finally, a few example puzzles generated by our algorithm are also provided. 

\end{abstract}

\section {Introduction}

Since their introduction in 1913 by Wynne \cite{bib:7}, crossword puzzles have maintained their popularity, and are a major linguistic game, as they are available in many languages. The puzzle consists of a grid of black and white squares. The objective of the puzzle is to fill the white squares with letters forming words or phrases, that are separated by black squares, by solving clues. 

Crossword puzzle presents two basic problems: (1) crossword generation, (2) crossword solving. Several effective solutions have been developed for crossword solving \cite{ernandes2005webcrow, bib:10,littman2002probabilistic, shazeer1999solving} where given a crossword grid and a set of clues, the task is to generate the solution to the puzzle. Crossword construction is a difficult and complex art, and the puzzles are mainly constructed by human experts called crossword constructors. As noted by Steinberg  and Last \cite{bib:1}, crossword construction typically starts with experts selecting a few theme words and designing the grid. Next, softwares such as Crossword Compiler \cite{bib:8}, Crossfire \cite{bib:9} or Phil \cite{bib:phil} are frequently used to speed up the task by filling up the rest of the grid with words from a database.

The problem of crossword puzzle construction involves two main steps: (1) selecting a set of words and clues, (2) given a set of words, fitting the words in a grid while respecting the constraints of the crossword. The first problem has been the focus of various studies, where efforts have been made to generate cohesive and thematic grids by using NLP based techniques \cite{bib:5} or using lexical relations encoded in databases \cite{bib:3}. The problem of grid filling has been posed as a constraint satisfaction problem \cite{bib:22} and also as a heuristic-based search problem \cite{bib:4}. Since crossword construction has been shown to be an NP-complete problem \cite{bib:11}, there is a need for effective heuristics/strategies to prune the search space. A few domain-independent crossword generation heuristics such as the cheapest-first heuristic, connectivity, and Smith’s adjacency restrictions have been discussed in the report on lessons learned by M. L. Ginsberg et al. \cite{bib:4}. 

Most of the earlier approaches \cite{bib:22, bib:14, bib:4, bib:21, bib:15} have focused on constrained crossword generation, where the puzzle geometry is assumed to be given. For a good
fill, dictionary richness
was pointed out to be an important factor in the performance of such approaches 
by Mazlack \cite{bib:20}. Harris et al. \cite{bib:18} remark that while there is scope for improving the efficiency of solving the constrained crossword puzzle generation (CCPG) problem, the basic problem is more or less solved. The success of CCPG can be observed from the success of crossword making softwares such as Crossword Compiler \cite{bib:8}, Crossfire \cite{bib:9}, and Phil \cite{bib:phil}. A brief overview of the research in CCPG can also be found in \cite{bib:21}.

However, there are applications that require handpicked words, such as themed puzzles, where the size of the vocabulary is limited. In such cases, the puzzle geometry needs to be determined as part of the puzzle generation. This is because a small vocabulary may not have a solution for a given fixed puzzle geometry. This unconstrained crossword puzzle generation problem (UCPG), as identified by Harris \cite{bib:17}, adds another layer of complexity as we not only need to determine the word locations, but also the position of the black squares in the layout. Harris \cite{bib:17} and Harris et al. \cite{bib:18} observed that UCPG is not only a generalization of the CCPG (since we can generate all possible puzzle geometries for an m x n grid and use the constrained puzzle algorithms to generate the solution set for each puzzle geometry), it also massively increases the complexity of the problem. 

\subsection {Approaches to the UCPG problem}

There is little work in the area of automation of UCPG that we are aware of. An algorithm that uses recursion with backtracking has been discussed by Harris \cite{bib:17}. However, their
puzzle size and input vocabulary size was small.
An automated generation of a variant of the unconstrained crossword puzzle called Crozzle, has been formulated in terms of {\em basic blocks} by Harris et al. \cite{bib:18}. A basic block is a geometric combination of words into a block, such that the removal of any word from the vocabulary corresponding to the grid produces an invalid crossword. Construction of basic blocks has been proposed as part of the strategies for automated solution of unconstrained crosswords. It must be noted that though a basic block is a valid crossword, not all unconstrained solutions are basic blocks. 

Another variant of the UCPG problem, called finding the diagram of diagramless crossword puzzles has been considered by Pershits and Stansifer \cite{bib:19}. They note that since the UCPG problem is ``hopelessly intractable" to solve even for average-sized puzzles, they create a  ``manageable" variant, in which, the words across and down, the size of the puzzle, and the order in which the answers appear in the grid are taken as input from which the puzzle is constructed. While the results show that the approach proposed is practical in finding the puzzle diagram, it does not solve the UCPG problem which needs to generate puzzles directly from hand-picked words. 

\subsection{Our contribution}

As observed by crossword constructors \cite{bib:111, bib:12}, even today, human constructors are seen  doing better than automated constructors, which keeps the problem open. The aim of our work is to introduce a set of new techniques to automate the process of crossword puzzle generation from a selected
set of words keeping the grid geometry open to accomodate "good" choices that are already picked up in the selected set.
We come up with general-purpose strategies that help in improving grid placements to generate good quality grids. Once the grid is generated, the work of refining and twisting the clues can be done by the human crossword constructor. We show that several algorithmic strategies can be combined to achieve a significant improvement in generation, and in a much fewer iterations. 

Summarily, in this paper, we propose a set of techniques for automatic solutions to the UCPG problem. A combination of strategies developed in this paper has been demonstrated to be successful in generating well-packed, unconstrained crossword puzzles, and our solution is also scalable w.r.t. the count of words that need to be fitted, wherein we observe a linear increase in computation. We also automate the process of clue generation, leading to end-to-end automation with a reduced intervention of human experts. We demonstrate the utility of the strategies devised through several examples and empirical analysis.

The paper is organized as follows. In the next section, we begin with an example puzzle generated by our approach. Next, in Section 3, we discuss the strategic functions used for the automated puzzle generation, which are referred to in the rest of the paper. Section 4 outlines the end-to-end algorithm which combines the strategies. In Section 5, we outline the main strategies. We describe them subsequently in detail and also talk about the effect of word sequence permutation order on grid fitting in Sections 6 - 10. Finally, in Section 11 we discuss the overall performance of the strategies, and also add a few crossword examples of various sizes that were generated from the algorithm. 

\section {An Example Puzzle Generation}

Crossword puzzle generation involves several optimizations including word selection, grid generation, alignment, movements, and fitting the boundaries. We formulate various strategies for automating these tasks. In addition, we also generate the clues automatically through a clue generation strategy.

\begin{figure}
\centering
\includegraphics[width=0.7\textwidth]{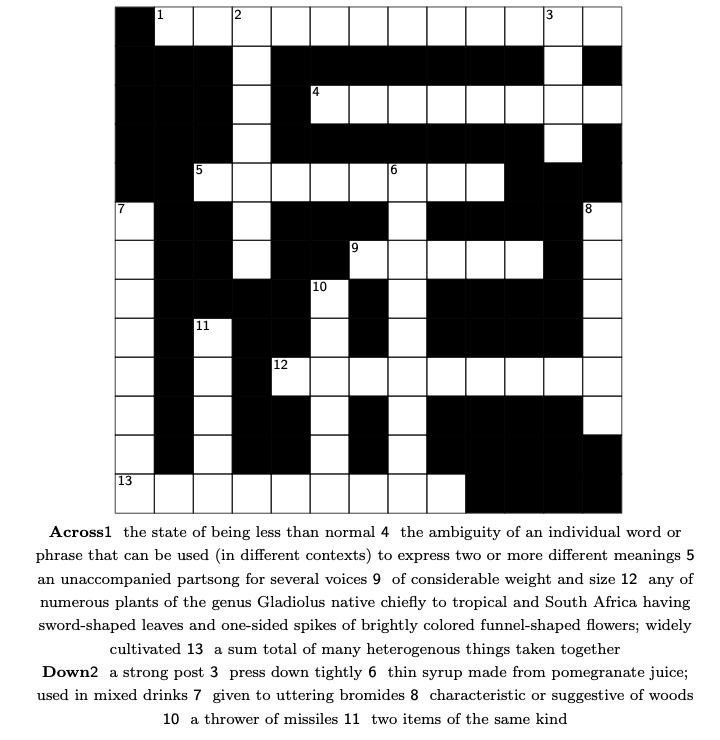}
\caption{Sample puzzle automatically generated using our algorithm.}
\label{fig:1}
\end{figure}

Consider for example the puzzle given in Fig \ref{fig:1}. It uses 13 words all in all, out of which 6 are aligned horizontally and 7 are aligned vertically in a 13 x 13 grid. This puzzle, including its clues, has been generated by our algorithm using the strategies developed in this paper. The solution for this puzzle is given in the appendix.

\section {Strategic Functions }

In this section, basic strategic functions that we use throughout the paper as part of the proposed strategies are described. 

\noindent 

G: Grid of words \\
L: Sorted list of words ranked with some ranking strategy. The list does not have duplicates.\\

\begin{enumerate}

\item {\bf Obtain the next fitting word}

\textit{pickWord (G, L):} Returns w, the first word in L, of which, the letters intersect with one or more words contained in G. First word in the list is a ``good'' pick if L is sorted on ranks by a ranking function suitable for grid fitting. If the list is unranked, the best ranked word needs to be returned by applying ranking at the time of the next pick.

 For example, 

  pickWord ([\textit{pick} (horizontal at 1,1 in grid), \textit{push} (vertical at 1,1 in grid)], [\textit{hall}, \textit{fresh}, \textit{bush}, \textit{lazy}]) 

returns \textit{hall} since `h’ intersects with \textit{push}.

\item {\bf Check for connectivity}

\textit{placeable (w, G)}: returns true if w can be added into G with at least one letter intersection with a word in G. This ensures connectivity of w in G. For example,

 placeable (\textit{hello}, [\textit{blue} (horizontal at 1,1 in grid), \textit{brown} (vertical at 1,1 in grid)]) 

returns {\bf false} since \textit{hello} does not intersect with any word in the grid.

\item {\bf Insert at intersection}

\textit{place (w, G)}: selects an orientation and places w in G, provided that \textit{placeable(w, G)} is {\bf true}, which makes sure that the new word w stays connected to at least one word in G. The new grid after this placement is returned. For example,

 place (\textit{red}, [\textit{two} (horizontal at 1,1 in grid), \textit{three} (vertical at 1,1 in grid)]) 

places \textit{red} at 3, 1 horizontally in G and returns the modified grid

\item {\bf Greedy removal strategy}

\textit{victim (G, L)}: returns a word u from G such that \textit{placeable(w, G \textbackslash u)} is true for some w in L. If no such word exists, it returns null. Essentially, this function returns a word that can be deleted from the grid, and due to which, one or more words from L can be accommodated in G. This function is used when there is no placeable word from the list for a given grid arrangement. The function then finds a victim word in the grid which if removed, allows the grid to accommodate at least one other word from the remaining list. This is used as a backtracking strategy, which may make it possible to grow the grid again after a word deletion. For example,

  victim ([\textit{apple} (horizontal at 2,1 in grid), \textit{cat} (vertical at 1,1 in grid)], [\textit{sink}, \textit{ant,} \textit{pink}, \textit{bush}]) 

returns \textit{apple} since \textit{ant} can be placed at 2,1 after removing \textit{apple} from the grid.

\item {\bf Last In First Out} 

\textit{wraparound (G, L):} removes the last added word from G and appends it to the tail of L and returns the new G and the new L. For example,

  wraparound ([\textit{four} (horizontal at 1,1 in grid), \textit{five} (vertical at 1,1 in grid)], [\textit{one, two, three}]) 

returns new grid [\textit{four} (horizontal at 1,1 in grid)], and new list [\textit{one, two, three, five}] assuming that \textit{five} was the last word added to G.

This is another backtracking strategic function besides \textit{victim()}. While \textit{victim()} is a more intelligent backtracking strategy, it may not always be able to find a victim satisfying our condition. To handle such cases, the LIFO wraparound strategy can be used repetitively till progress can be made. 

\end{enumerate}

\section{ Algorithm \textit{CrosswordGenerator}}

The algorithm moves through the following stages:

\begin{enumerate}

\item {\bf Seed and Initialization:} The algorithm begins with picking a set of seed words (S) that are to be fit in the grid. The grid size is initialized to dimension N X N such that N is the length of the largest word in the list, since we need at least a grid that is at least large enough to fit each word in the list. The grid that starts eventually changes and adjusts its dimensions as more words get placed one by one. Thus, the initial grid dimension is a function of the seed words.

\item {\bf Ranking:} It generates a distance matrix for a given word list, by computing the distance of every word with every other word in the list, and then uses it to rank words based on the number of intersections that they have with other words. Once the seed words are ranked and the list is sorted according to the ranks in ascending order, we begin with placement. We discuss the effect of ascending-first and descending-first choices later in Section 6 when we discuss the behavior of the distance used in ranking. 

\item {\bf Placement:} The placement begins with placing the first word at the center of the grid. Then, we repeatedly call \textit{pickWord} (G, L) to find the next word which can be placed. If a word can be placed in the grid, we place it using \textit{place (w, G)}. If no such word exists, we invoke strategic function \textit{victim} (G, L) to remove a word from the grid which allows the addition of more words. If no victim can be found, then we call strategic function \textit{wraparound} (G, L) to remove the most recently added word. The process is repeated until no more words can be added to the grid.

\item {\bf Termination condition:} At any point, the best grid (in terms of the number of words added) found so far is stored and the placement is repeated with a larger grid. The algorithm can be terminated based on various criteria such as the size of the grid, time limit, desired grid density (utilization), or the number of words added, etc.

\end{enumerate}

\begin{algorithm}
\KwData{\\
S: List of words \\
G: Grid \\
N: Grid size (N X N grid) \\
R: The Rank vector for words which stores the ranks of words corresponding to order given in S. The words are ranked based on commonalities with other words \\
L: sorted list of ranked words, a temporary variable \\
T: upper bound on the number of iterations for refinement \\
i: Iteration count \\
$G_{best}$: best grid found so far in terms of count of the number of words inserted \\
$n_{best}$: no. of words in $G_{best}$ \\
}

{\bf Initialize:} \\
vocabulary S to the list of seed words (about a couple of dozens) \\
Initialize N, the grid size as the size of the largest word in S \\
Initialize $n_{best}$ to 0 \\
Initialize i, iteration count to 0 \\
Initialize G to the empty grid. \\
Initialize $n_{words}$ to 0. \\
Initialize grid epoch $GE_{MAX}$ = 500. We reset a grid if it cannot be handled in $GE_{MAX}$ iterations. \\
{\bf Ranking:} \\
Compute R, the rank vector assigning rank to every word \\
Find the distance of each word with every other word \\
Rank of a word is the sum of its distances  \\
$L \gets \text{sort S, based on ranks in R}$ \\
{\bf Placement and Termination: }\\
Ltop = first word in L \\

\Repeat{L is not empty and $i < T$}{
\If{$n_{words}$ == 0}{Place $L_0$, the first word from L at the center of G (across or down). \\
Remove $L_0$ from L}
\Else{
w = pickWord (G, L) \\
\If{w is not nil}{
G = place (w, G) \\
$n_{words}$ ++ \\
If $n_{words} > n_{best}$: \\
$G_{best} \gets G$ \\
$n_{best} \gets n_{words} $\\}
\Else{
let v = victim (G, L) \\
\If{v is nil}
{G, L = wraparound (G, L)}}} 
i ++ \\
\If{i \% $GE_{MAX}$ == 0 or L.head == Ltop (back to the same list)}{
N = N + 1 \\
Reset grid with dim N x N \\
$n_{words}$ = 0 \\
$L \gets \text{sort S, based on R}$ // Reset words }}
output $G_{best}$ as the current best solution found \\
\caption{Algorithm to automatically generate a crossword.}
\end{algorithm}

\section{Strategies}

This section outlines several strategies for puzzle generation into four classes of strategies, which are, {\em ranking and selection of words}, {\em word placement}, {\em grid resizing and grid reshaping} and {\em clue generation}. The strategies are then described and analyzed in detail in subsequent sections. 

\begin{enumerate}

\item {\bf Ranking and Selection of words:} The strategy for ranking words is at the heart of our approach, as it constrains the inflow of words into the grid, influencing the placement of the subsequent words on the grid. The words are qualified through selection as per the order determined by the ranking method. 

\item {\bf Word placement:} The placement strategy decides how the words are placed on the grid, and more importantly, what to do when no more words can be placed. This is handled through backtracking policies, without overriding the ranking order. There are two ways that our algorithm incorporates as word placement strategies, which are finding a {\em victim word} for replacement, and {\em wrap around} to withdraw words one by one from the grid. 

\item {\bf Grid resizing and reshaping:} The size of the grid is an important constraint, and the goal is to find a dense grid with more words in a given grid size. When no more words can be placed in the grid even with word placement strategies, this iterative strategy of grid resizing is used. Also, grid reshaping can allow redistribution of the grid space so that new areas can become free for adding more words.

\item {\bf Clue generation:} The strategy for clue generation is formulated to automatically generate clues for the fitted words. Clues may be generated from a resource such as the dictionary. The clues can be either taken {\em as it is} or can be further refined and {\em twisted} by a human expert.

\end{enumerate}

\section{Ranking and Selection}

In this section, we first show how choice of words based on ranking can make a difference to the performance, and then systematically develop the ranking strategies.

\subsection{Effect of word ordering on no. of words fitted}

\begin{figure*}
	\begin{subfigure}{0.5\textwidth}
		\includegraphics[width=\textwidth]{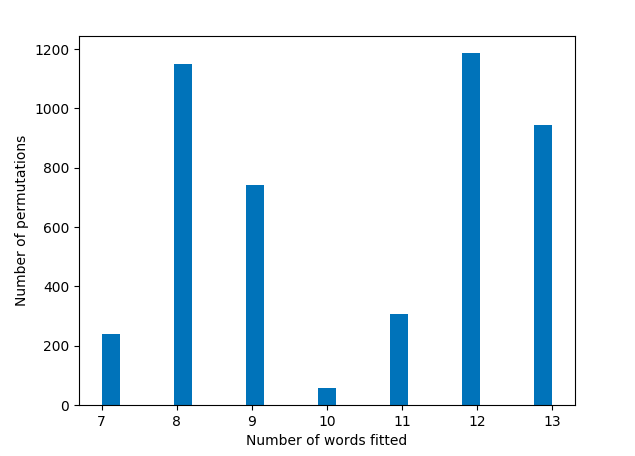}
		\caption{No. of permutations = 4625}
	\end{subfigure}
	\begin{subfigure}{0.5\textwidth}
		\includegraphics[width=\textwidth]{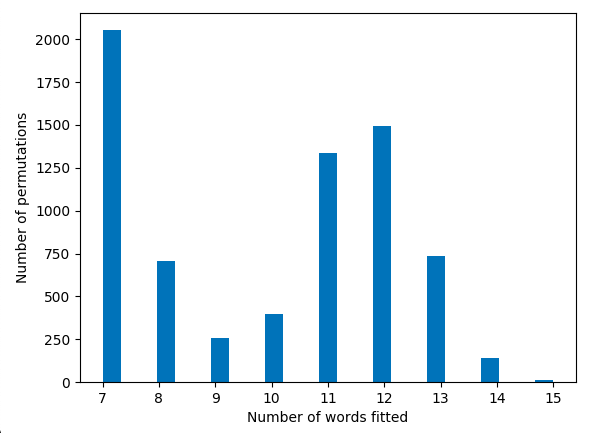}
		\caption{No. of permutations = 7135}
	\end{subfigure}
	\begin{subfigure}{0.5\textwidth}
		\includegraphics[width=\textwidth]{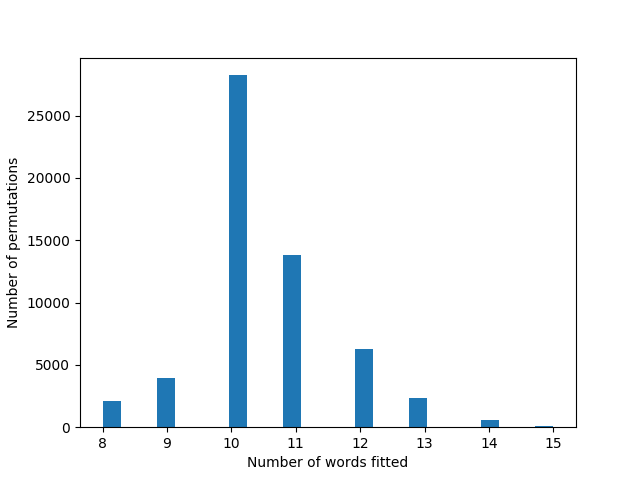}
		\caption{No. of permutations = 57325}
	\end{subfigure}
	\begin{subfigure}{0.5\textwidth}
		\includegraphics[width=1\textwidth]{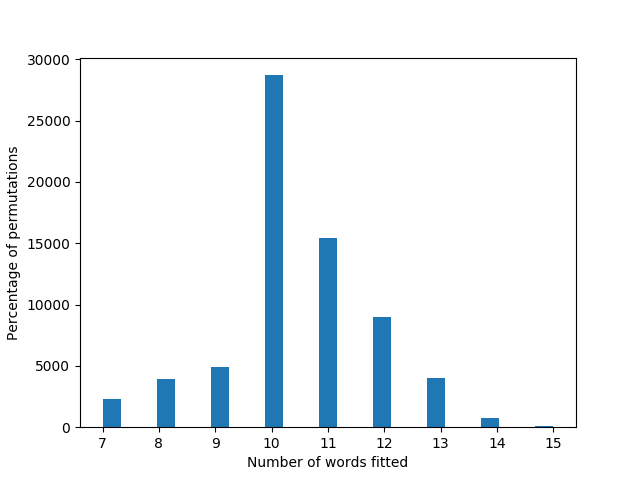}
		\caption{Total number of permutations = 69082 (Aggregate)}
		\label{fig:2d}
	\end{subfigure}
	\begin{subfigure}{0.5\textwidth}
		\includegraphics[width=1\textwidth]{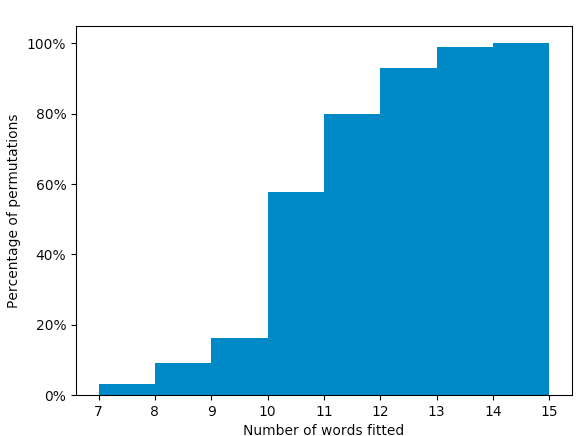}
		\caption{Cumulative Histogram for the aggregate data}
		\label{fig:2e}
	\end{subfigure}
	\caption{Effect of number of words fitted in a 13x13 grid based on the order of the words for the brute force algorithm. The dataset consists of 26 words and permutations have been randomly sampled from the possible 26! = 4.03e26 possible permutations.}
	\label{fig:2}
\end{figure*}

Given an input set of words, the order in which they are placed into the grid matters. We have found that certain sequences work much better than others. This subsection explores how the sequence ordering of words affects the performance of the grid. In this subsection, we do not use any backtracking or heuristic to improve the performance, but primarily bring out the problem, and then go on to describe strategies to improve the performance of grid creation. Plots in Fig. \ref{fig:2} show the number of words fitted for a randomly chosen set of permutations without backtracking as per the following brute force strategy:

\noindent \textbf{for every permutation of the word list:}

 \indent  \textbf{try placing words one by one till no word fits}

 \indent  \textbf{store the grid when no more word can be fitted}

From Fig. \ref{fig:2} (a)-(d), we can see that as we increase size of the set of permutations, the curve becomes a single bell curve showing normal distribution of permutations over no. of words that can be fitted without backtracking. From Fig \ref{fig:2e} we can observe that approximately 80\% of the permutations were stuck below 12 words, and 60\% were stuck below 11 words, as compared to the best we found in this test (17 words). This indicates that a heuristic based on word selection can be useful to reorder the words through back tracking strategies to increase the capacity of the grid, and in fewer iterations avoiding brute force search through permutations. Next, we develop ranking strategies analyzing their impact on grid performance, starting with an intersection based ranking strategy.

\subsection{Intersection based Ranking}

Let us consider a list of seeds words, \{WROTE, BREAD, LOBBY, HELLO\}. Table \ref{tab:1} shows the intersection matrix (shown full) for the list. Each entry in the table represents the count of the number of letters common to the corresponding words. The last column shows the total number of intersections that each word has with other words in the entire list. 

\begin{table}
\caption{Example Intersection Matrix}
\begin{tabularx}{\textwidth}{|
p{\dimexpr 0.19\linewidth-2\tabcolsep-2\arrayrulewidth}|
p{\dimexpr 0.13\linewidth-2\tabcolsep-\arrayrulewidth}|
p{\dimexpr 0.13\linewidth-2\tabcolsep-\arrayrulewidth}|
p{\dimexpr 0.13\linewidth-2\tabcolsep-\arrayrulewidth}|
p{\dimexpr 0.12\linewidth-2\tabcolsep-\arrayrulewidth}|
p{\dimexpr 0.3\linewidth-2\tabcolsep-\arrayrulewidth}|} \hline 
 & WROTE & BREAD & LOBBY & HELLO & Number of intersections \\\hline 
WROTE & - & 2 & 1 & 2 & 5 \\
BREAD & 2 & - & 0 & 1 & 3 \\ 
LOBBY & 1 & 0 & - & 2 & 3 \\
HELLO & 2 & 1 & 2 & - & 5 \\
\hline 
\end{tabularx}
\label{tab:1}
\end{table}
From this matrix, we can generate the ranking as a sequence (WROTE, HELLO, BREAD, LOBBY) with ties resolved arbitrarily. The selection of words during insertion into the grid is carried out in the order of the ranking. There can be other variants to rank the words, e.g., ranking based on the number of words intersected. 

\subsubsection { Effectiveness of Intersection based Ranking}

\begin{figure}
\centering
\includegraphics[width=0.8\textwidth]{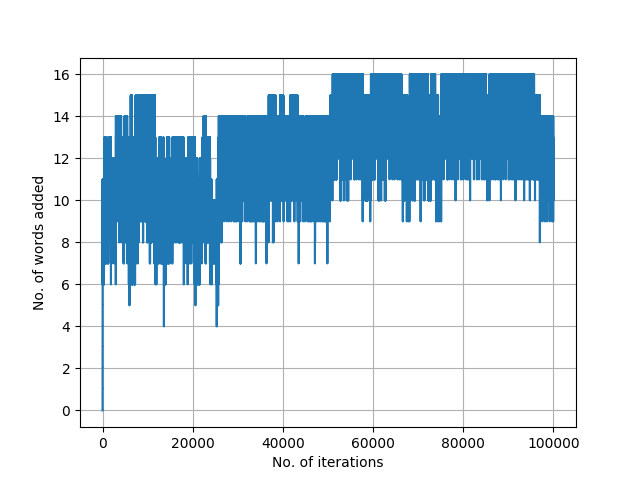}
\caption{Assigning order on the basis of number of intersections with other words (ascending)}
\label{fig:3}
\end{figure}

\begin{figure}
\centering
\includegraphics[width=0.8\textwidth]{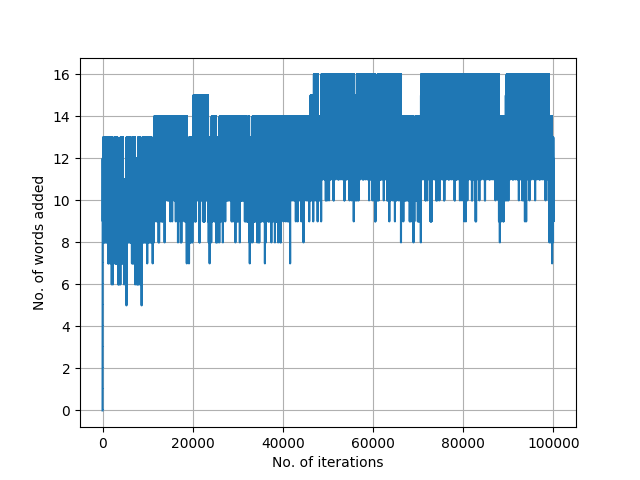}
\caption{Assigning order on the basis of number of intersections with other words (descending)}
\label{fig:4}
\end{figure}

To start with, Figures \ref{fig:3} and \ref{fig:4} show the effect of selection done on the basis of the number of intersections discussed above. Backtracking step is taken whenever no more words can be added using this ranking. For these figures, the grid size is kept constant. As the number of iterations increase, more and more words get added, till the grid is saturated. Backtracking changes the puzzle by swapping words. In this test, last in first out backtracking was adopted. In this particular example, we can observe that at about 50,000 iterations, the grid gets saturated. The figures also explore if there is any effect of ranking order on grid saturation. Figure \ref{fig:3} chooses words in ascending order of ranking, and Figure \ref{fig:4} chooses words in descending order of ranking. One can observe that the saturation point does not depend on the ranking order when the ranking is a function of the number of intersections. Thus we can conclude that in this case, the ranking function is not very effective since changing the order of ranking has not had much effect on the grid saturation point. We significantly improve this saturation point in the following subsection, which introduces a strategy based on a new ranking metric called {\em pebble and sand} metric.

\subsection{Pebble and sand strategy for Ranking}

A change in this ranking function improves the above result considerably. In the pebble and sand strategy, the following {\em pebble and sand ranking} function $R$ is formulated. Here, instead of considering letter level intersection, word level intersections are used. As we shall see through our analysis, this choice significantly improves the performance.

Given word w in word list l, \\

\begin{equation*}
R(w, l) = \frac{\text{no. of words in l having intersection with w}}{\text{total no. of words in l}}
\end{equation*}

Here, the ranking is independent of grid content and it represents a {\em futuristic} choice. At a given point of insertion, it can be noted that the content of the grid is already chosen as per their ranks. So, it makes sense to consider this option of ignoring the content of the grid while ranking words for the next choice, since the words were already chosen as per their ranks in the list.

\begin{figure}
\centering
\includegraphics[width=0.8\textwidth]{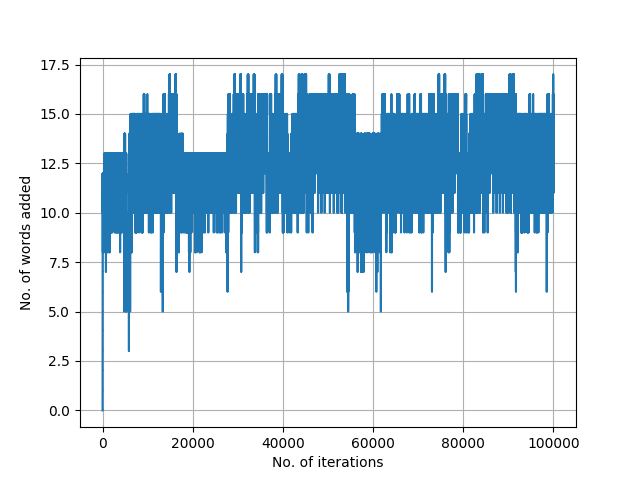}
\caption{Assigning the order based on the number of words intersected (ascending)}
\label{fig:5}
\end{figure}

\begin{figure}
\centering
\includegraphics[width=0.8\textwidth]{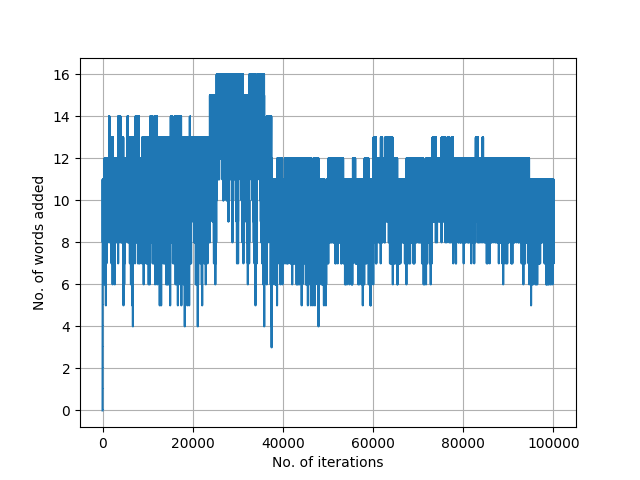}
\caption{Assigning the order based on the number of words intersected (descending)}
\label{fig:6}
\end{figure}

We can observe through Fig. \ref{fig:5} and \ref{fig:6} that when the ranking is thus based on the number of words intersected and when the words are ordered in ascending order, the grid saturates at much fewer iterations. It saturates much faster at about 18000 iterations in ascending order and about 24000 iterations in descending order, as compared to about 50000 in the letters based intersection based strategy. Also, the saturation point is 17 words and 16 words for ascending and descending orders respectively. 

The descending order does not beat the ascending order as shown in Fig. \ref{fig:6}. This indicates that it is better to place the words that are apart from each other first. This strategy can be compared with strategy used for physically filling pebbles and sands in a jar. In our case, {\em sand} represents words that have more intersections with other words, and {\em pebbles} are words that have fewer. By following the ascending order the pebbles are filled first, whereas, the descending order fills the sand first limiting future choices considerably. So, filling pebbles first opens up more future choices allowing the grid to grow. 

\begin{figure}
\centering
\includegraphics[width=0.8\textwidth]{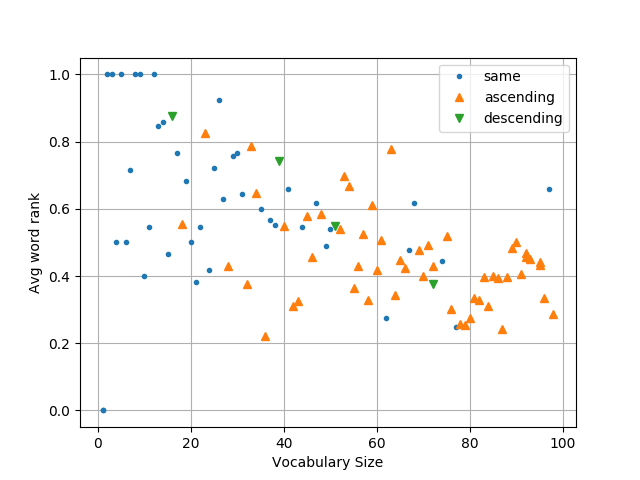}
\caption{Effect of Size}
\label{fig:7}
\end{figure}

To see the effect of vocabulary size and average word rank on the pebble and sand strategy, we considered 100 randomly chosen vocabulary sets of varying sizes from size 1 to 100, with one dataset per vocabulary size. The grid size was constant. Fig. \ref{fig:7} shows the results. We observe that when the vocabulary size is small, the ranking order does not matter (circle points). However, when the vocabulary size increases, the pebble and sand strategy works better (triangle-up points) than descending order. The triangle-down points indicate where the descending order works better. Out of 100 data sets, 4 triangle-down points, 54 triangle-up points and 42 circle points were obtained. This establishes the effectiveness of ascending order, i.e. the pebble and sand strategy.

\section {Towards Word Placement Strategies}

Pebble and sand ranking metric rates a word in terms of what it shares with the rest of the words in the input vocabulary. Based on the pebble and sand idea, now we define a distance between a permutation and the best ranked (pebble and sand strategy order) permutation. This {\em pebble and sand distance} of a permutation is computed as follows:

Given best ranked permutation $P_{best}$ and the worst ranked permutation is $P_{worst}$ (flipped order of $P_{best}$), the distance a permutation $P$ is defined as:

\begin{equation}
D(P) = \frac{ \sum_{i=0}^{N} | pos(w_i, P) - pos(w_i, P_{best}) | }{\sum_{i=0}^{N} | pos(w_i, P_{worst}) - pos(w_i, P_{best}) | } 
\end{equation}

where $pos(w,p)$ gives position of word $w$ in permutation $p$, $N$ is the no. of words in $P$, which is same as no. of words in $P_{best}$.

For example D([ab, sh, io]) w.r.t permutation [sh, io, ab] is $(2+1+1) / 4$, i.e., it is 100\% away from permutation [sh, io, ab]. Similarly, D([sh, ab, io]) w.r.t permutation [sh, io, ab] is 50\% and D([ab, io, sh]) w.r.t permutation [sh, io, ab] is 100\%.

\begin{figure}
\centering
\includegraphics[width=0.7\textwidth]{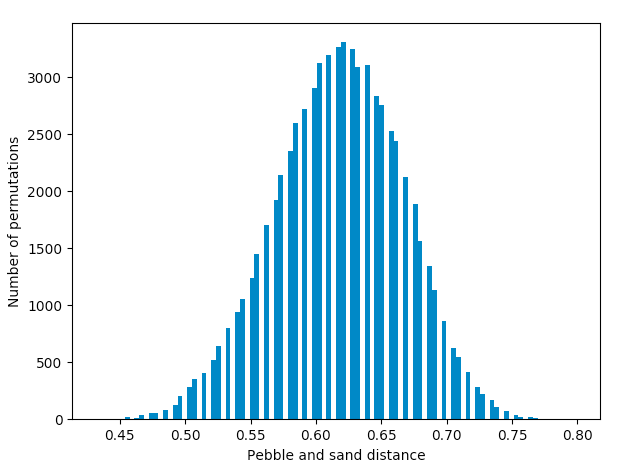}
\caption{Distribution of pebble and sand distance over permutations.}
\label{fig:8}
\end{figure}

\begin{figure}
\centering
\includegraphics[width=0.7\textwidth]{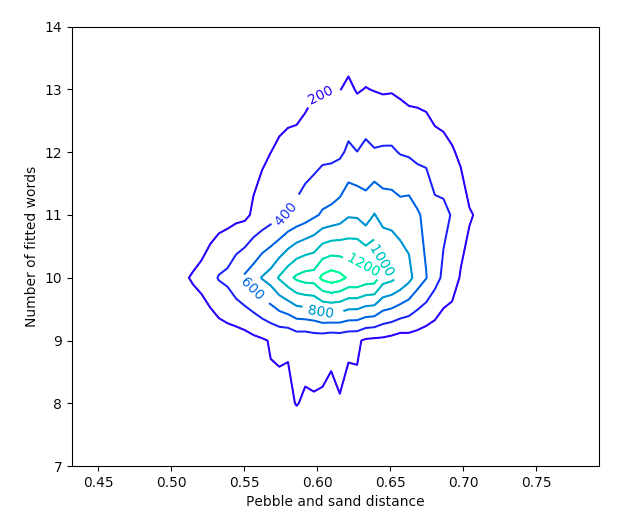}
\caption{Effect of pebble and sand distance on the number of words fitted without backtracking. The contour plot contains 75.3 \% of the data points.}
\label{fig:9}
\end{figure}

Figure \ref{fig:8} shows the distribution of the pebble and sand distance over a total of 69,082 permutations. The dataset consists of 26 words and permutation sequences over those words have been randomly sampled from the possible 26! (4.03e26) possible permutations. It can be observed that the distance is normally distributed over the chosen random set of permutations. The distribution follows a bell curve. We have seen in Fig. \ref{fig:2d} that the number of permutations are normally distributed over no. of words fitted without backtracking. The combined effect is shown in the contour plot given in Fig. \ref{fig:9}, which shows a hill like behavior. The peak in Figure \ref{fig:9} shows the distance at which the highest number of permutations occur, which is close to 0.6. The goal of our algorithm is not to achieve this peak, but to shift it, which is rather to shift the whole contour towards the upward area and make it steeper. This is so that no matter how far away a given permutation is from the best permutation, we can show that one can still have a “good” strategy that can quickly place a higher number of words in a given grid. The contours are revisited and the improvement is shown when we discuss the performance of our word placement strategies.


\section{Word Placement Strategies}

\begin{figure}
\centering
\includegraphics[width=0.4\textwidth]{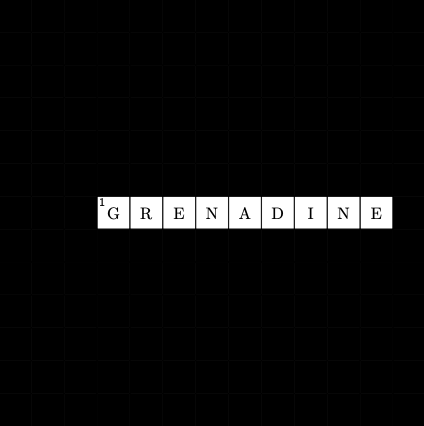}
\hspace{0.01\textwidth}
\includegraphics[width=0.4\textwidth]{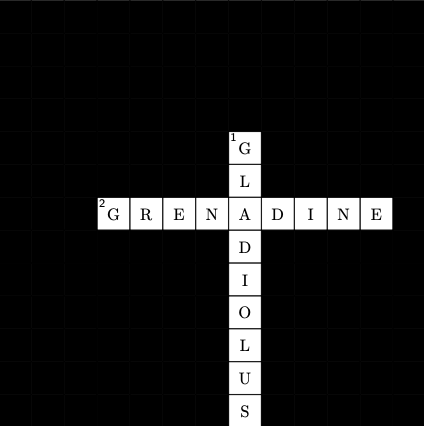}
\vspace{0.01\textwidth}
\includegraphics[width=0.4\textwidth]{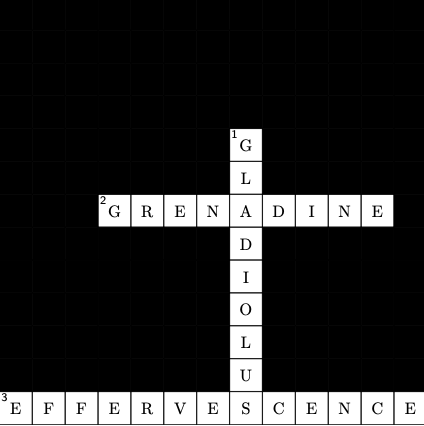}
\caption{Illustration of word placement}
\label{fig:10}
\end{figure}

We now illustrate the strategies for placing words. To initialize, the chosen word is placed at the center. Next, from the list, we iteratively choose further words that have at least one intersection with words already in the grid. Fig. \ref{fig:10} given below shows the step by step process of building up the grid. We start with the best-ranked word, ‘GRENADINE’. Next, ‘GLADIOLUS’ intersects with ‘GRENADINE’ at ‘A’, and hence it can be placed top down. Next, the word ‘EFFERVESCENCE’ gets placed across. In this manner, the grid gets built up piecewise till this process gets interuppted. We now discuss two placement strategies to handle a situation when no word from the input list can be added to the grid.


\subsection { Last In First Out}

\begin{figure}
\centering
\includegraphics[width=0.4\textwidth]{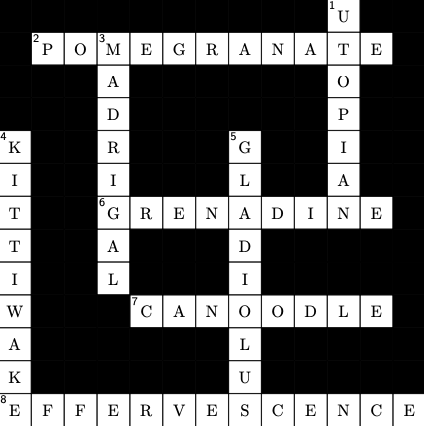}
\hspace{0.01\textwidth}
\includegraphics[width=0.4\textwidth]{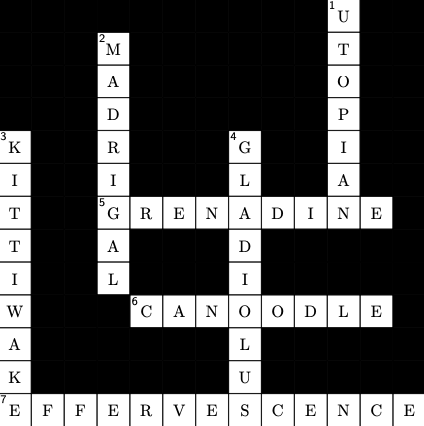}
\caption{Illustration of backtracking}
\label{fig:11}
\end{figure}

In this strategy, we remove the most recently added words one by one till another word can be added to the grid. Removed words are pushed back into the input vocabulary list at the end. It can be observed that in the left image in Fig. \ref{fig:11}, no more words can be added. Hence, we remove the last added word, ‘POMEGRANATE’ from the grid. This strategy is helpful as it can avoid deadends through backtracking.

\begin{figure}
\centering
\includegraphics[width=0.8\textwidth]{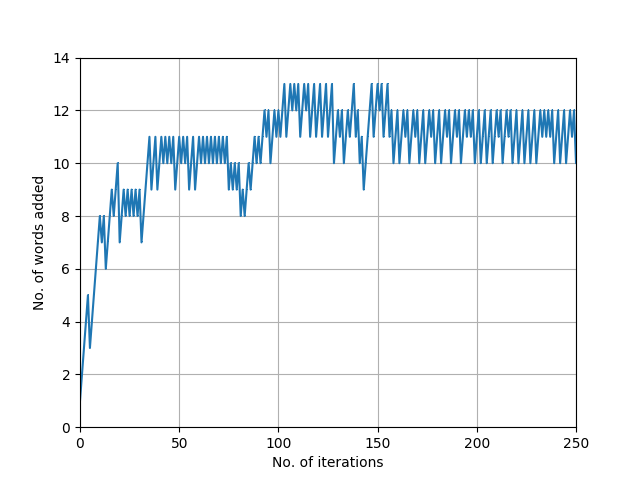}
\caption{Last In First Out Strategy}
\label{fig:12}
\end{figure}
As shown in Figure \ref{fig:12}, the LIFO strategy reaches ten words in about 25 iterations exploring all words in the wordset chosen (26 words in this case). In the graph, every downward edge shows a backtracking step using the LIFO strategy. Over 250 iterations, the graph shows that a maximum of 13 words were reached with this strategy. Each point in this graph is a valid crossword. We improve the result further through the greedy removal strategy explained below.

\subsection {Greedy Removal Strategy}

\begin{figure}
\centering
\includegraphics[width=0.332\textwidth]{images/image37.png}
\hspace{0.01\textwidth}
\includegraphics[width=0.332\textwidth]{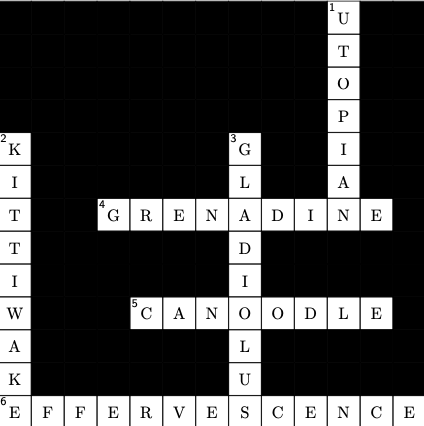}
\hspace{0.01\textwidth}\\
\includegraphics[width=0.332\textwidth]{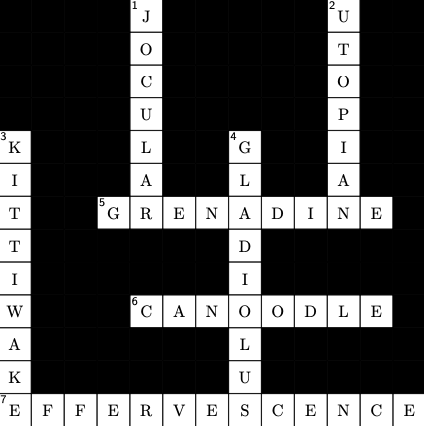}
\caption{Illustration of victim}
\label{fig:13}
\end{figure}

In this strategy, using the function \textit{victim (G, L)} described in Section 3, we find a word, such that removing the word allows new words to be placed in the grid. For example, in the first image in Fig. \ref{fig:13}, no new words can be added. Hence, we choose to remove the word ‘MADRIGAL’ as a search over the input list finds the word ‘JOCULAR’ that can fit in. The second figure shows the grid after removing the victim word. The third figure shows the grid with one new word ‘JOCULAR’ added. 

\noindent

{\bf Post-processing for Disconnections:} Note that this strategy can result in the grid being disconnected, and post processing may be needed to reconnect the components of the grid if disconnected words are not desired. This post-processing consists of connecting the disconnected components by finding connecting words through strategies such as pattern search used by Meehan and Gray \cite{bib:13}. In their approach, the grid layout is fixed and regex-based pattern matching is used to find words fitting in a fixed layout. Our set of strategies allow us to proceed with grid construction without fixing the layout. However, for post-processing, when the layout is more or less stabilized, the regex-based strategy can be used for connecting disconnected components.

\begin{figure}
\centering
\includegraphics[width=0.8\textwidth]{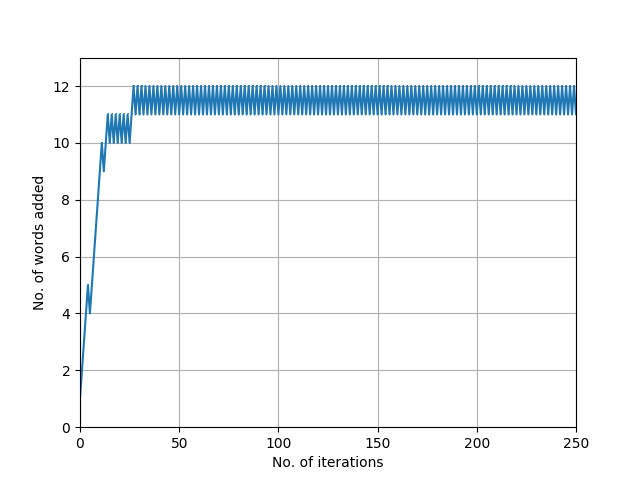}
\caption{Greedy Removal Strategy (Victim) saturated in a cycle}
\label{fig:14}
\end{figure}
Fig. \ref{fig:14} shows the result of the greedy removal strategy that is implemented without using the LIFO word placement strategy explained in the section above. The saturation point is achieved much quicker, here, in less than 50 iterations. The saturation capacity is also higher than the one achieved with LIFO in the same number of iterations. 

\noindent

{\bf Preventing Cyclic Removals:} We can also observe that greedy removal has eventually saturated in a cycle of two words causing cyclic \textit{removal and addition} showing flicker around 12 and 11 words. In some cases, the cycle length in terms of no. of words may be bigger. In such cases, greedy removal finds a victim word from the grid for replacement, and in a cyclic removal, the same set of words get chosen. Suitable cycle prevention strategies may be used. If datasets are small, cycles are more probable, especially when many of the words in the dataset are already in the grid, leaving a much smaller set of words for replacement. Cyclic replacements can be handled by reducing priorities to words once backtracked, and bringing them back eventually if needed. Similar strategy is used in tabu search \cite{bib:16}, in which, a finite list of most recent nodes (tabu words, in our case) is maintained. The tabu tenure count (tt) is defined as the size of the tabu list. Similarly, once a word has been deleted from the grid, it can only be added back after tt other moves are completed, meaning all other words have been given a chance. 

\subsection{Effectiveness of Word Placement Strategies}

\begin{figure}
\centering
\includegraphics[width=0.8\textwidth]{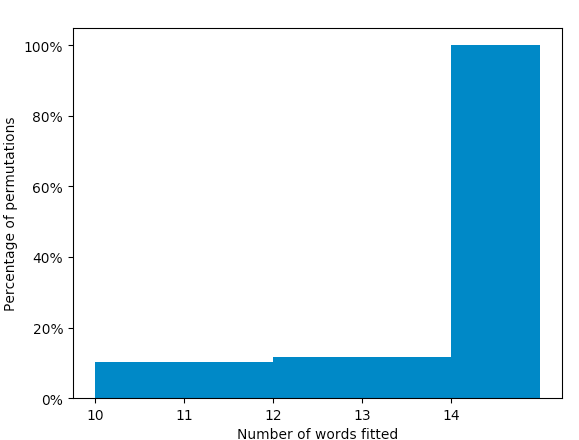}
\caption{Effect of number of words fitted in a 13x13 grid based on the order of the words using backtracking strategies. The grid size is kept constant (13 x 13) and each permutation is used to fit the words over 50 iterations.}
\label{fig:15}
\end{figure}

Fig. \ref{fig:15} shows the distribution of the permutations with the number of words fitted using the two word placement strategies presented above, but without using ranking and grid resizing. For the sake of comparison, the grid size is kept constant (same as the grid size considered in Fig \ref{fig:2e}). As we can observe, about 90\% of the permutations are able to fit 14 words within just double the number of iterations in brute force word placement. 

Fig. \ref{fig:15} represents the performance with our word placement backtracking strategies without using ranking. When the figure is compared with Fig. \ref{fig:2e}, which represents the brute force placement without ranking or backtracking, we can observe that the distribution has now shifted significantly towards the right side, indicating that with our strategies, most of the permutations (over 90\%) were able to fit 14 words. Also, this result was achieved within 50 iterations, which is double the number as compared to 26 iterations taken in Fig. \ref{fig:2e}. The reader is reminded that the input data size in terms of no. of words is 26.

\begin{figure*}
	\begin{subfigure}{0.5\textwidth}
		\includegraphics[width=\textwidth]{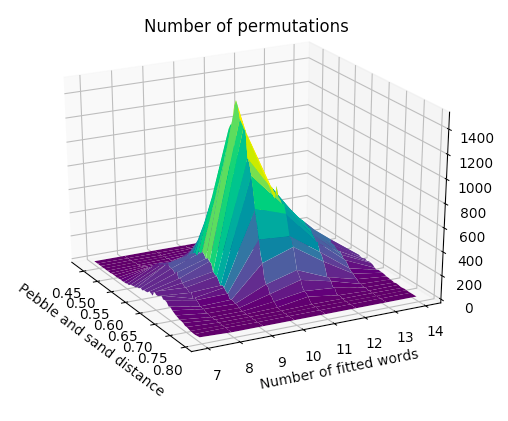}
		\caption{Using brute force word placement}
		\label{fig:16a}
	\end{subfigure}
	\begin{subfigure}{0.5\textwidth}
		\includegraphics[width=\textwidth]{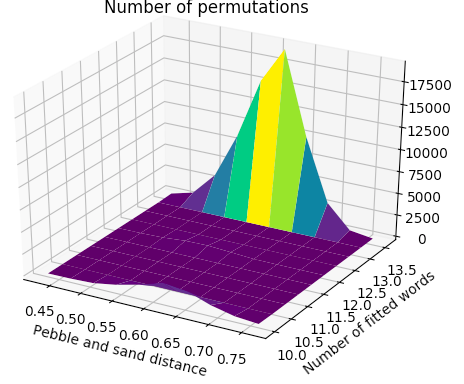}
		\caption{Using backtracking strategies}
		\label{fig:16b}
	\end{subfigure}
	\caption{Effect of pebble and sand distance on the number of words fitted.}
	\label{fig:16}
\end{figure*}

Fig \ref{fig:16b} shows the 3-d surface for the number of permutations fitted for a given number of words fitted vs. its pebble and sand distance, when our backtracking strategies were used. For comparison, similar surface is plotted for the above brute force strategy, which is shown in Fig. \ref{fig:16a}. We can observe that with backtracking, the peak has not only become sharper, but it also gets shifted towards a higher number of words to be fitted. It may be noted that the contour plot in Fig. \ref{fig:9} and Fig. \ref{fig:16a} are for the same data. 

\section{Grid resizing and reshaping}

\begin{figure}
\centerline{
\includegraphics[width=0.33\textwidth]{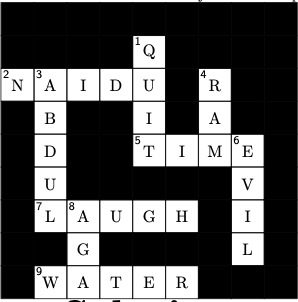}
\hspace{0.001\textwidth}
\includegraphics[width=0.33\textwidth]{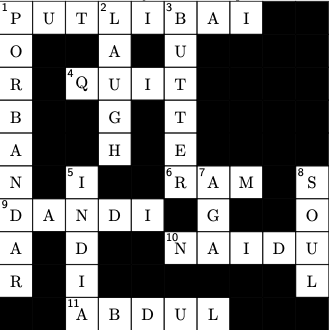}
\hspace{0.001\textwidth}
\includegraphics[width=0.33\textwidth]{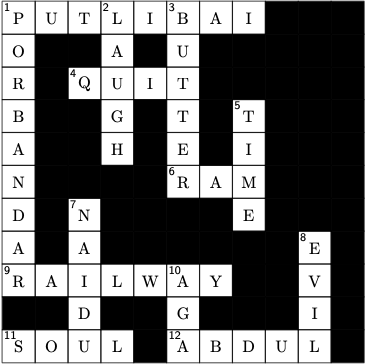}}
\vspace{5mm}
\centerline {\hspace{0.001\textwidth}
\includegraphics[width=0.36\textwidth]{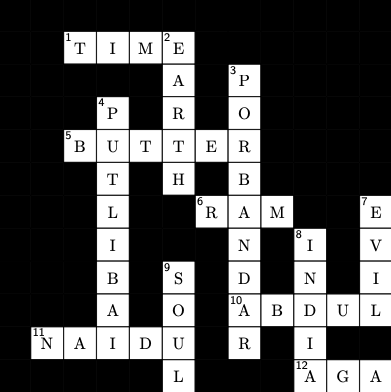}
\hspace{0.001\textwidth}
\includegraphics[width=0.36\textwidth]{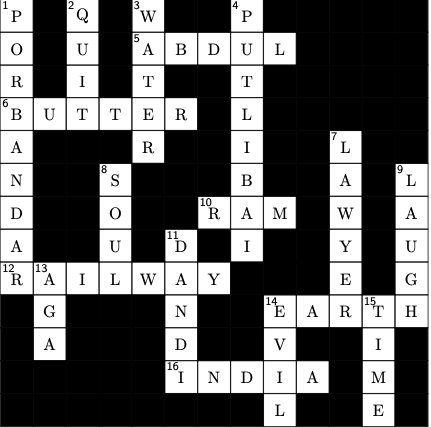}
}
\caption{Illustration of grid resizing}
\label{fig:17}
\end{figure}

In this section, we discuss strategies for changing the {\em high-level structure} of the grid for a better placement. We start with a grid of the size of the largest word and try new placements. Next, we iteratively increase the size of the grid, towards achieving the {\em desired utilization} of the grid space. One way to define a desired utilization is to go for the size of the grid that allows maximum placement of words. Figure \ref{fig:17} shows the placement of a given set of words on different grid sizes. The smallest grid in the figure is of size 9 x 9, which allows 9 words to be fitted from the given vocabulary following the ranking order, and within some time limit. In the same conditions, the largest grid of size 13 x 13 allows 17 words to be placed. 

\begin{figure}
\centering
\includegraphics[width=0.4\textwidth]{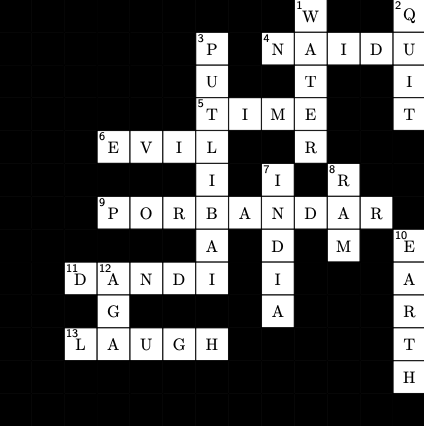}
\hspace{0.01\textwidth}
\includegraphics[width=0.4\textwidth]{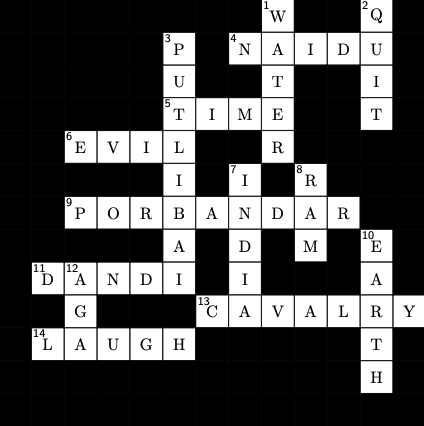}
\caption{Illustration of grid reshaping/repacking}
\label{fig:18}
\end{figure}

Another strategy is to repack the grid. This allows redistributing the mass of the grid equally. For example, if one part of the grid is mostly sparse, while the other side of the grid is densely filled, we can rebalance the grid by removing an empty row (or column) from one side and adding the row (or column) to the other side. The left image in Fig. \ref{fig:18} shows the grid with the left side being sparse, and the right image shows the grid after the first column is removed from the left side and added to the right side. As we can see, one new word (CAVALRY) can be placed after this rebalancing.

Thus, the placements are heuristics-based in terms of our ranking, placement and grid resizing strategies, and a solution acceptable to human expert can be chosen as the algorithm continues to generate a sequence of solutions by applying the placement strategies. At any point of time the algorithm can be terminated and the best solution found till then can be accessed. 

\section{Clue Generation}

For clue generation, we refer to an English dictionary to look up word definitions. The definitions and meanings of words can be used to generate the clues. For example, the dictionary definition \cite{bib:23} of ‘ABYSMAL’ is ‘immeasurably low or wretched : extremely poor or bad’. This is a simple and useful approach as word definitions do not contain the target word, nevertheless, conveying the meaning of the word. Once we have the clues, they can be manually altered. Similarly, word usages may also be used after removing the word and filling it with blanks. For example, for the same word, a clue \cite{bib:23} can be ``They were living in {\ldots}{\ldots}{\ldots}. ignorance’’.

\section{Overall Performance}

Further to the results on effectiveness of the strategies in their respective sections, in this section, we now discuss the overall performance of the algorithm, and provide empirical results and also a few example puzzles generated through our implementation of the algorithm. First the scalability of the algorithm w.r.t. the number of words to be fitted is demonstrated. Next, we discuss the quality of packing of the grids in terms of the share of the number of black squares. Finally, we provide a few example crossword puzzles generated using our algorithm.

\subsection{Scalability of the algorithm}

\begin{figure}{t}
\centering
\includegraphics[width=0.8\textwidth]{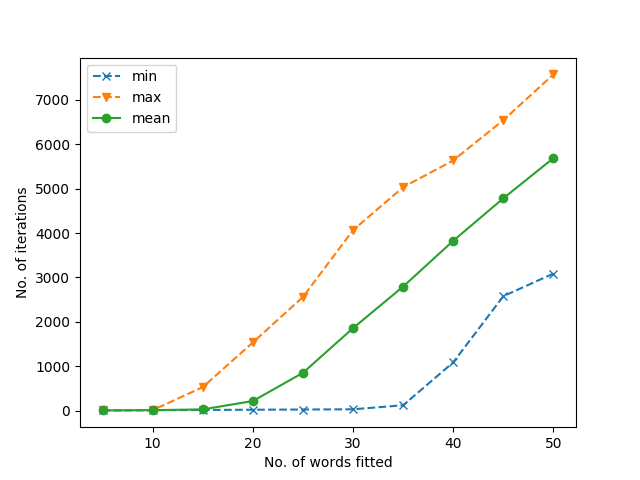}
\caption{Number of iterations vs. number of words fitted with grid resetting on expansion}
\label{fig:19}
\end{figure}

Combining the above strategies (ranking, victim words, LIFO backtracking, grid resizing) as per the algorithm outlined in Section 4, we obtain the results as shown in Fig \ref{fig:19}. It shows the maximum, minimum, and the average number of iterations taken to fill the grid in the range 5 - 50 words, where the vocabulary size is chosen to be 100. A back-tracking or a word fitting step is counted as a single iteration. As the grid size gets initialized to the size of the largest word in the vocabulary, we can see from the figure that fitting upto 15 words takes much fewer no. of iterations, since there are not many conflicts as the grid can accommodate those many words. 

The selection of words is based on order defined by the pebble and sand ranking strategy. Initial grids are sparse grids. However, as more words need to be fitted, more iterations are needed, and word placement backtracking strategies are put to use. As shown in the figure, with added iterations, the no. of words fitted increased following a near linear rise. The results are averaged over 100 datasets consisting of 100 words each, chosen randomly from the English dictionary. The grid has been reset on increasing the grid size after every round (epoch) of 500 iterations. When the grid resizing (expansion) step takes place, the grid is first reset to initialization. For example, if 6000 iterations are taken, then the grid undergoes a dozen resets, and it is also expanded that many times by one column and one row each time. This is an {\em iterative refinement strategy}. The refinement can be terminated when a {\em desired} result is obtained as discussed earlier.

\begin{figure}
\centering
\includegraphics[width=0.8\textwidth]{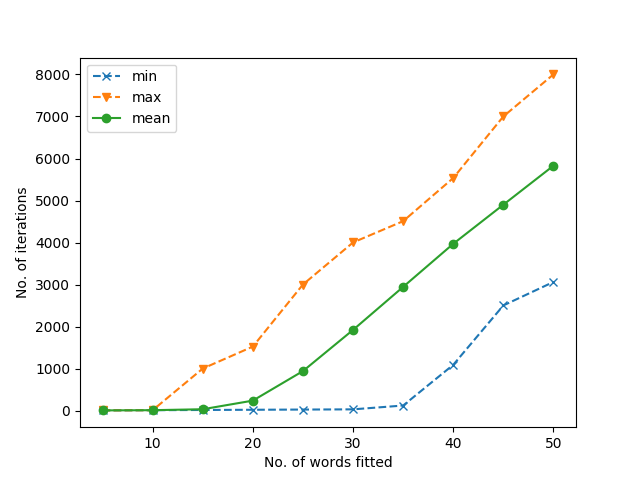}
\caption{Number of iterations vs. number of words fitted with grid continuation on expansion}
\label{fig:20}
\end{figure}

{\em Grid continuation} is another possibility to explore when a grid expansion takes place. However, grid continuation was not found to be improving results. In fact, the results with the {\em iterative refinement by grid reset on expansion} were a bit better in the average sense as shown by the green line in Fig. \ref{fig:19}, when compared to Fig. \ref{fig:20}, which shows the results with {\em grid continuation on expansion}. Also, the band between the max and the mean is smaller for grid continuation as compared to grid reset. This behavior indicates that {\em grid reset} brings more freedom right from the beginning as compared to {\em grid continuation}, which ties the grid to an already chosen grid that is settled for a smaller size. 

\subsection{On Grid Utilization}

 \begin{figure}
	\begin{subfigure}{0.5\textwidth}
		\includegraphics[width=\textwidth]{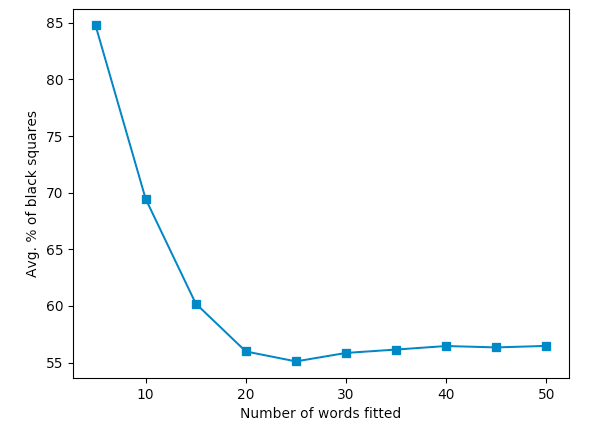}
		\caption{Without resetting }
		\label{fig:21a}
	\end{subfigure}
	\begin{subfigure}{0.5\textwidth}
		\includegraphics[width=\textwidth]{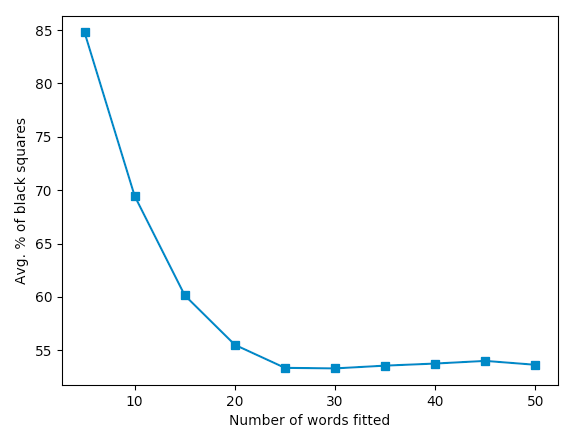}
		\caption{With resetting}
		\label{fig:21b}
	\end{subfigure}
	\caption{Average utilization of the grid w.r.t the number of words fitted}
\end{figure}

Figures \ref{fig:21a} and \ref{fig:21b} show the utilization of the grid w.r.t. the number of words fitted with and without using the grid reset strategy. As we can observe, for a need of a smaller number of words, the utilization is low, since the grid size is initially set to the largest word in the vocabulary, due to which, the grid was left sparse. However, as we increase the number of words that we want to be fitted in, the grid utilization, on an average, rises to around 45\% - 55\% leaving black squares to occupy 55\% - 45\% of the grid. The utilization is a little higher in the grid resetting strategy as compared to grid continuation, since every time a grid expansion takes place, the grid is started afresh using the iterative refinement technique opening up more options. The utilization results are comparable to those reported in \cite{bib:17}, which achieved 43.75\% of black squares with grid sizes 8 X 8 with a small vocabulary size. We achieved with the new techniques comparable numbers with much larger grids and on much larger input vocabularies in a limited number of iterations.

\begin{figure}
	\begin{subfigure}{0.5\textwidth}
		\includegraphics[width=\textwidth]{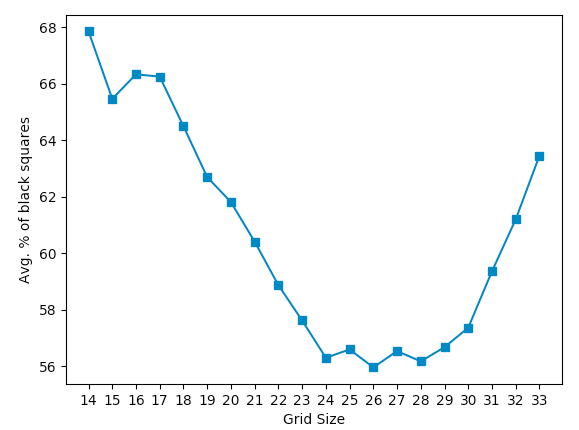}
		\caption{Without resetting }
		\label{fig:22a}
	\end{subfigure}
	\begin{subfigure}{0.5\textwidth}
		\includegraphics[width=\textwidth]{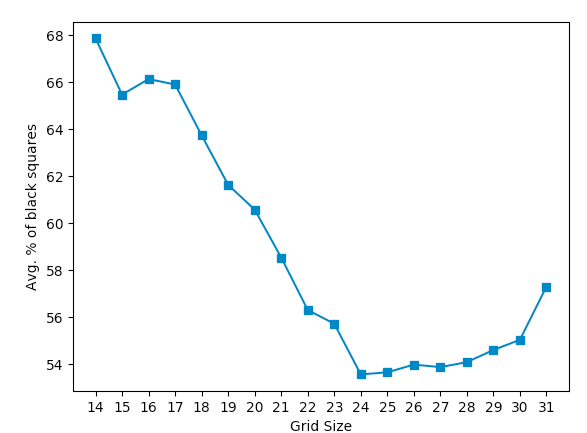}
		\caption{With resetting}
		\label{fig:22b}
	\end{subfigure}
	\caption{Average utilization of the grid w.r.t the grid size}
\end{figure}

Figures \ref{fig:22a} and \ref{fig:22b} show the utilization of the grid w.r.t. grid size for the same data. As we can observe, the best utilization is achieved in the range 24 x 24 to 28 x 28. Again, we find that grid resetting strategy results in better grids. The no. of iterations taken before freezing a puzzle ranged from 5 to 8000, depending on the number of words that were required to be fitted. A rise at the end of the graphs is observed for higher grid sizes, since the maximum number of iterations was kept limited. 

\subsection{Crossword Samples}

\begin{figure}
\centering
\includegraphics[width=0.6\textwidth]{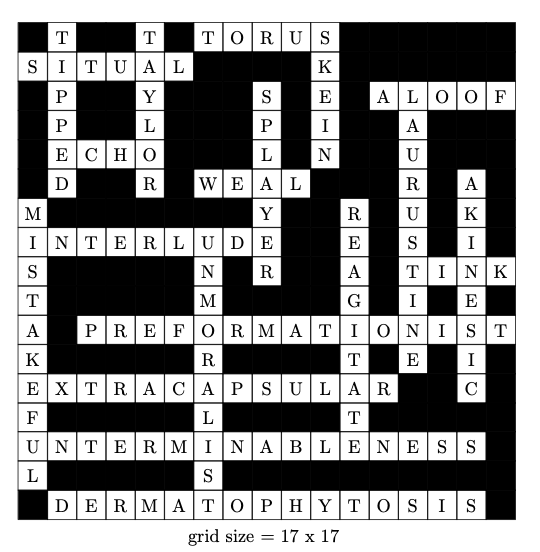}
\caption{Puzzle with 20 words}
\label{fig:23}
\end{figure}

\begin{figure}
\centering
\includegraphics[width=0.6\textwidth]{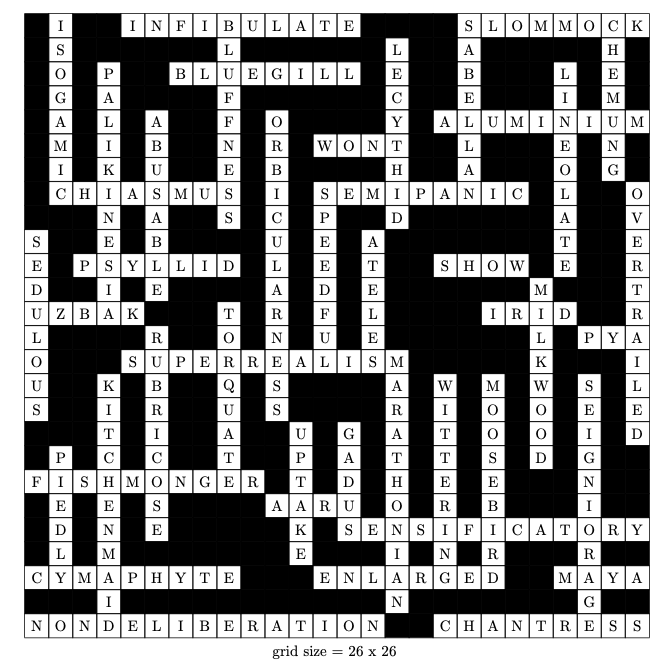}
\caption{Puzzle with 45 words}
\label{fig:24}
\end{figure}

\begin{figure}
\centering
\includegraphics[width=0.6\textwidth]{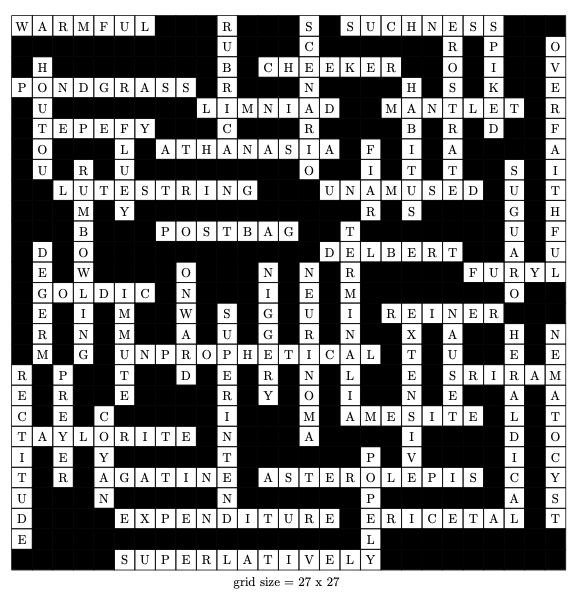}
\caption{Puzzle with 50 words}
\label{fig:25}
\end{figure}

\begin{figure}
\centering
\includegraphics[width=0.7\textwidth]{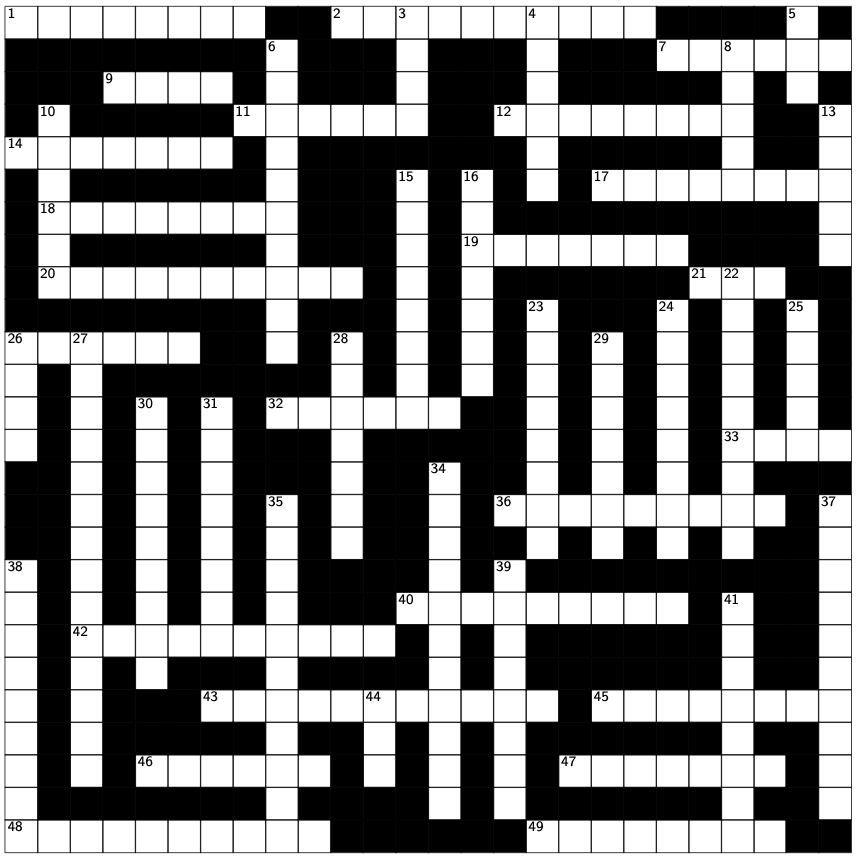}
\\
\includegraphics[width=0.6\textwidth]{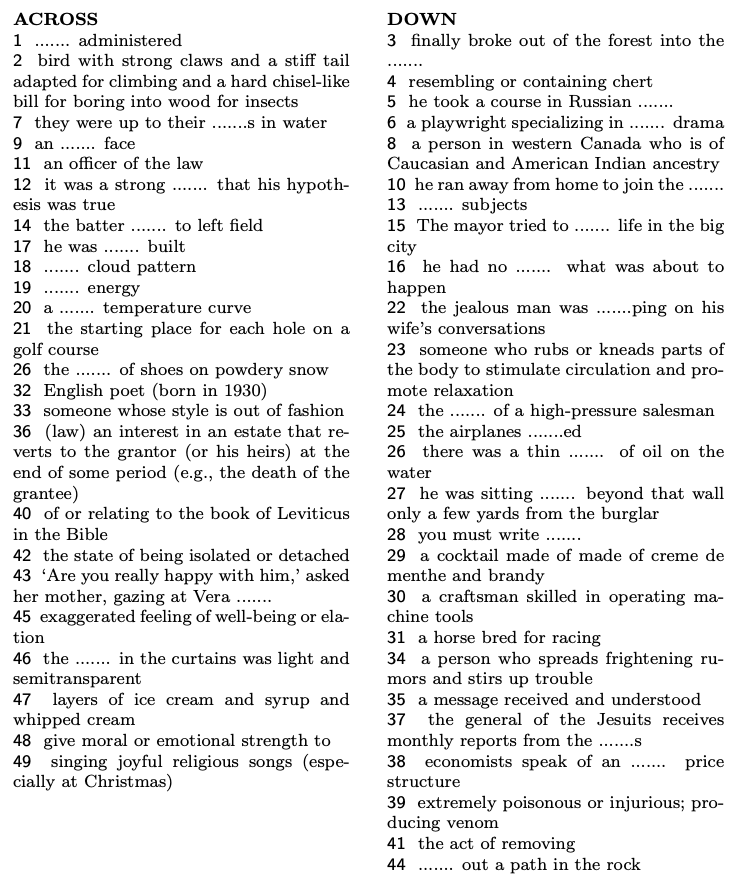}
 
\caption{Sample Unsolved Puzzle}
\label{fig:26}
\end{figure}

Figures \ref{fig:23}, \ref{fig:24} and \ref{fig:25} show three example crosswords (solutions) generated using the combined strategy. They consumed 66 (for 20 words), 5563 (for 45 words), and 4948 (for 50 words) iterations respectively. Figure \ref{fig:26} gives a complete crossword puzzle with clues which was automatically generated using our algorithm. The clues were generated using definitions in the Python-based TextBlob library \cite{bib:24} and usages in the WordNet lexical database \cite{bib:25} supported in Python-based NLTK library \cite{bib:nltk}. The solution to the puzzle in Figure \ref{fig:26} can be found in the appendix. 

\section{Conclusion}

We developed and demonstrated a new algorithm for end-to-end unconstrained crossword puzzle generation as a combination of several strategies. The algorithm works on a given set of words and dynamically constructs a grid from it by expanding and adjusting it as and when needed. It combines several strategies covering word selection, word placement, grid design. It also includes clue generation stage using a resource such as a dictionary. An important advantage of our algorithm is that it can work with a considerably small set of vocabulary, and our experiments show that the algorithm is able to fit a significant subset of the words from a small set in a well-packed grid. This was revealed through empirical analysis over 100 data sets of size 100 words each. In each data set, words were randomly selected from a dictionary. In all cases, we were able to generate puzzles of sizes up to 50 words iterating under 7000 to 8000 iterations. The algorithm therefore can be used in applications where words need to be handpicked, such as in thematic puzzles. 

We explored the effect of reordering of vocabulary word sequences and found that certain sequences work better than others. To reach closer to better performing sequences, we developed (i) a ranking metric for efficient selection of words and (ii) a couple of backtracking strategies. The {\em pebble and sand strategy} has been found to perform well in most cases. It significantly brings down the grid saturation point in terms of the number of iterations, at the same time increasing the number of words fitted. A combination of LIFO and greedy removal strategies makes the word placement less sensitive to the ordering, and as observed in the empirical results, it significantly increases the number of words fitted within much fewer iterations. We have shown that our algorithm also works well with a large number of words, with only a linear rise in the number of iterations. Further, we have observed that it is better to reset a grid when doing grid expansion than to continue with the same grid, as the same grid constrains the new search space. The clue generation strategy provides a basic clue set for the crossword, which can be altered manually to make them more interesting.

These strategies are tunable, and hence they can be modified and adapted for use in other word games. Future work areas include ranking functions, victim selection, grid packing strategies, and explorations into properties of vocabularies in terms of their abilities to generate better quality puzzles.

\bibliographystyle{apa}
\bibliography{sample}

\section*{Appendix: Solutions to Puzzles}
\label{appendix:a}
\begin{figure}[h!]
\centering
\includegraphics[width=0.7\textwidth]{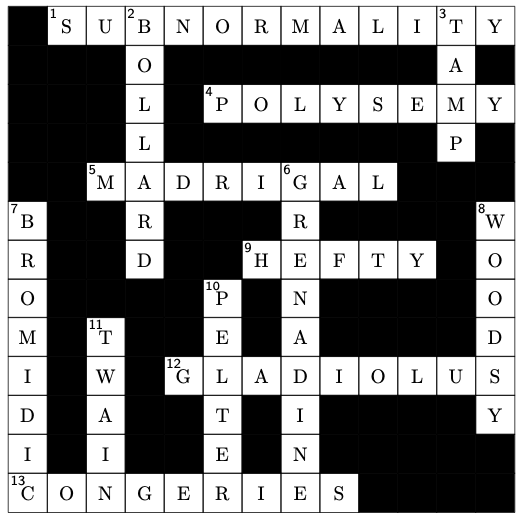}
\caption{Solution of sample puzzle given in Figure \ref{fig:1}.}
\label{fig:27}
\end{figure}

\begin{figure}[h!]
\centering
\includegraphics[width=1.0\textwidth]{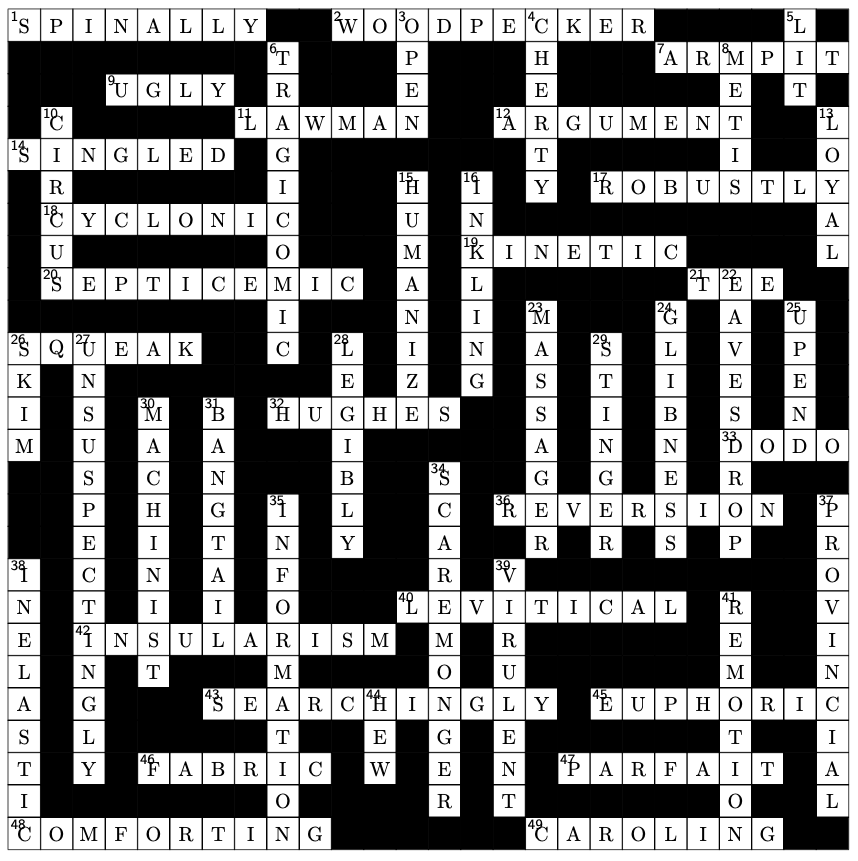}
\caption{Solution of sample puzzle given in Figure \ref{fig:26}.}
\label{fig:28}
\end{figure}

Figure \ref{fig:27} gives the solution to the puzzle given in Figure \ref{fig:1}. Figure \ref{fig:28} gives the solution to the puzzle given in Figure \ref{fig:26}. 

\end{document}